\documentclass{article}

\PassOptionsToPackage{numbers}{natbib}
\usepackage[preprint]{neurips_2021}

\usepackage[utf8]{inputenc} % allow utf-8 input
\usepackage[T1]{fontenc}    % use 8-bit T1 fonts
\usepackage{hyperref}       % hyperlinks
\usepackage{url}            % simple URL typesetting
\usepackage{booktabs}       % professional-quality tables
\usepackage{amsfonts}       % blackboard math symbols
\usepackage{nicefrac}       % compact symbols for 1/2, etc.
\usepackage{microtype}      % microtypography
\usepackage{algorithm}
\usepackage{algorithmic}
\usepackage{graphicx}
\usepackage{wrapfig}

\usepackage{bm}
\usepackage{amssymb}
\usepackage{xcolor}
\usepackage{amsmath}
\usepackage{amsthm}

\newtheorem{proposition}{Proposition}

\title{SoftDICE for Imitation Learning:\\ Rethinking Off-policy Distribution Matching}

\author{%
  Mingfei Sun\thanks{Corresponding author: \texttt{mingfei.sun@cs.ox.ac.uk}} \\
  University of Oxford\\
  \And
  Anuj Mahajan \\
  University of Oxford\\
  \And
  Katja Hofmann  \\
  Microsoft Research\\
  \And
  Shimon Whiteson \\
  University of Oxford\\
}

\begin{document}

\maketitle
\begin{abstract}
We present SoftDICE, which achieves state-of-the-art performance for imitation learning.
SoftDICE fixes several key problems in ValueDICE~\cite{kostrikov2019imitation}, an off-policy distribution matching approach for sample-efficient imitation learning.
Specifically, the objective of ValueDICE contains logarithms and exponentials of expectations, 
for which the mini-batch gradient estimate is always biased.
Second, ValueDICE regularizes the objective with replay buffer samples when expert demonstrations are limited in number, which however changes the original distribution matching problem. 
Third, the re-parametrization trick used to derive the off-policy objective relies on an implicit assumption that rarely holds in training. 
We leverage a novel formulation of distribution matching and consider an entropy-regularized off-policy objective, which yields a completely offline algorithm called SoftDICE. 
Our empirical results show that SoftDICE recovers the expert policy with only one demonstration trajectory and no further on-policy/off-policy samples.
SoftDICE also stably outperforms ValueDICE and other baselines in terms of sample efficiency on Mujoco benchmark tasks.
\end{abstract}

\section{Introduction}
The recent success of reinforcement learning (RL) in many domains~\cite{mnih2015human,silver2014deterministic} showcases the great potential of applying this family of learning methods to real-world applications. 
A key prerequisite for RL is the ability to design a reward function that specifies what agent behavior is preferable. 
However, in many real-world applications, designing a reward function is prohibitively difficult, as it requires balancing many competing factors.
For example, when training a physical robot to navigate, it can be challenging to design a reward function that properly characterizes human-like navigation. 
By contrast, demonstrations are often readily available in such real-world applications.
Imitation learning studies exactly this problem -- given a limited number of expert demonstrations, learn an optimal  policy without any further access to the expert policy or reinforcement signals of any kind.

% distribution matching; on-policy vs off-policy
Recently, distribution matching has gained popularity in imitation learning and shown state-of-the-art performance in many benchmark tasks. 
Specifically, distribution matching approaches interpret state-action pairs provided in the expert demonstrations as sample points from a target distribution. 
Imitation learning is then framed as learning a policy that minimizes a divergence between this target distribution and the distribution induced by the policy.
The optimization process, reminiscent of Generative Adversarial Nets~\cite{goodfellow2014generative}, 
often consists of two steps~\cite{ho2016generative,ghasemipour2020divergence,wang2019random}: first, estimate the density ratio of state-action pairs between the target distribution and the learned policy; second, use these ratios as reinforcement signals in standard RL algorithms to update the learned policy.
The main limitation of this formulation is that estimating those density ratios typically requires on-policy samples that are generated by rolling out the learned policy.
Consequently, every update to the policy requires new interactions with the environment from the learned policy, which precludes many real-world applications where interactions are expensive and limited. 
One recent study, ValueDICE~\cite{kostrikov2019imitation}, relaxes the on-policy constraint by using the change of variable trick~\cite{nachum2019dualdice}, which then yields a completely off-policy objective and a practical algorithm with state-of-the-art sample efficiency.

% value dice method
However, ValueDICE has several key problems that can affect its empirical performance and convergence guarantees.
% gradient estimate is biased
First, the off-policy objective in ValueDICE contains logarithms and exponentials of expectations (i.e., $\log$/$\exp$ applied to an expectation), which makes any mini-batch estimate of the gradient biased~\cite{belghazi2018mine,kostrikov2019imitation}.
% fail to learn expert distribution
Second, the objective actually adopted in the ValueDICE algorithm is different from the original off-policy one and thus provides no guarantee that the expert policy is learned. 
Third, the change of variable used to derive the off-policy objective implicitly assumes that the Markov chain induced by the learned policy is always ergodic, which rarely holds during training as the sampling may be terminated early by the environment reset.

In this paper, we propose a new objective based on the Earth-Mover Distance (EMD) that eliminates all the aforementioned problems while still remaining off-policy. 
Using EMD draws connections to the classic feature matching approaches in inverse reinforcement learning~\cite{abbeel2004apprenticeship,ratliff2006maximum,ziebart2008maximum} and the resulting intermediate objective can be interpreted as feature expectation matching (FEM) with respect to arbitrary learned features.
This connection also motivates the use of the principle of maximum entropy to regularize training when the number of demonstrations is extremely small. 
Moreover, the change of variable trick adopted in ValueDICE~\cite{kostrikov2019imitation} can be rigorously improved by a more practical and unbiased off-policy estimate.
The new resulting entropy-regularized objective naturally yields a practical algorithm, \textit{SoftDICE} that is completely offline and requires no further on-policy or off-policy samples in training. 
Our empirical results show that SoftDICE can successfully learn the expert policy with only one demonstration trajectory and no further interactions from the environment.
Furthermore, SoftDICE stably outperforms ValueDICE and other baselines in terms of sample efficiency across different continuous control tasks in the Mujoco benchmark environment. 

\section{Background}

\paragraph{Notation}
We consider an infinite-horizon Markov Decision process (MDP) with 
a finite state space $\mathcal{S}$, 
a finite action space $\mathcal{A}$, 
a transition kernel $p: \mathcal{S}\times\mathcal{A}\times\mathcal{S}\rightarrow[0, 1]$, 
a reward function $r: \mathcal{S}\times\mathcal{A}\rightarrow\mathbb{R}$ which is unknown, 
a discount factor $\gamma \in [0, 1)$, 
and an initial state distribution $p_0$ from which $s_0$ is sampled. 
$\Pi$ is the class of policies, $\pi: \mathcal{A}\times\mathcal{S}\rightarrow[0, 1]$ that the learner is considering.

%Any policy has a duality with its state-action distribution. 
For a policy $\pi\in\Pi$, define its state-action  distribution as  $d(s, a)=(1-\gamma)\sum_{t=0}^{\infty}\gamma^t d_t(s, a)$, 
where $d_t(s, a)=P(s_t=s, a_t=a|s_0\sim p_0, \forall i<t, a_i\sim\pi(\cdot|s_i), s_{i+1}\sim p(\cdot|s_i, a_i)$.
A basic result from~\cite{puterman2014markov} is that the set of valid state-action distribution $\Omega\triangleq\{d_{\pi}: \pi\in\Pi\}$ can be written as a feasible set of affine constraints:
if $p_0(s)$ is the distribution of starting states ($p_0(s)>0 \forall s\in\mathcal{S}$) and $p(s^\prime|s, a)$ is the transition model, then
\begin{equation}\label{equ:feasible-constraint}
\Omega=\big\{ d\geq 0: \int_{a}d(s, a)\text{d}a = (1-\gamma) p_0(s) + \gamma\int_{s^\prime, a^\prime}p(s|s^\prime, a^\prime)d(s^\prime, a^\prime)\text{d}s^\prime\text{d}a^\prime,  \quad \forall s\in\mathcal{S} \big\}. 
\end{equation}
Intuitively, this constraint states that a feasible state-action distribution is a stationary distribution induced by an ergodic (i.e., irreducible and aperiodic) Markov chain with transition probability $(1-\gamma)p_0(s) +\gamma p(s|s^\prime, a^\prime)$\footnote{We consider the Markov chain induced by simulating an MDP with discounted reward. }.
Furthermore, there is one-to-one correspondence between $\Pi$ and $\Omega$:
\begin{proposition}\label{prop:stationarity-constraint}
\cite{syed2007game} If $d(s, a)$ is feasible by~\eqref{equ:feasible-constraint}, then $d$ is the state-action distribution for $\pi(a|s) \triangleq d(s, a)/\sum_{a}d(s, a), \forall (s, a)$, and $\pi$ is the only policy whose state-action distribution is $d$. 
\end{proposition} 
Proposition~\ref{prop:stationarity-constraint} forms the foundation of a distribution matching approach for imitation learning that learns $\pi$ by minimizing the divergence between state-action distribution of the policy $d_{\pi}(s, a)$ and the empirical distribution $d_E(s, a)$ of state-action pairs in the demonstration~\cite{ho2016generative,ke2019imitation,kostrikov2019imitation,ghasemipour2020divergence}.

The problem formulation in ValueDICE is based on the KL-divergence. 
Specifically, it adopts the Donker-Varadhan representation~\cite{donsker1975asymptotic} of KL-divergence and formulates the objective as follows, 
\begin{equation}\label{equ:valuedice-kl-divergence}
\mathcal{D}_{\text{KL}} (d_\pi || d_E) = \max_{x:\mathcal{S}\times\mathcal{A}\rightarrow\mathbb{R}} -\log \mathbb{E}_{(s, a)\sim d_E}\big[ e^{x(s, a)} \big] + \mathbb{E}_{(s, a)\sim d_\pi}\big[ x(s, a) \big].
\end{equation}
To make the objective practical for off-policy learning, ValueDICE takes inspiration from derivations used in DualDICE~\cite{nachum2019dualdice}, and performs the following change of variable:
\begin{equation}\label{equ:valuedice-change-of-variable}
x(s, a) = \nu(s, a) - \mathcal{B}^{\pi}\nu(s, a), 
\end{equation}
where $\mathcal{B}^{\pi}$ is the expected Bellman operator with respect to policy $\pi$ and zero reward:
\begin{equation}\label{equ:change-of-variable-trick}
\mathcal{B}^{\pi}\nu(s, a) = \gamma\mathbb{E}_{s^\prime\sim p(\cdot|s, a), a^\prime\sim\pi(\cdot|s^\prime)}[\nu(s^\prime, a^\prime)]. 
\end{equation}
% Such change of variable is only valid for $\log\frac{d_{\pi}(s, a)}{d_E(s, a)}\in\mathcal{K} \quad \forall (s, a)\in\mathcal{S}\times\mathcal{A}$,  where $\mathcal{K}$ is a bounded subset of $\mathbb{R}$,  and $x$ is restricted to the family of functions $\mathcal{S}\times\mathcal{A}\rightarrow\mathcal{K}$. 
With this change of variable, the second expectation telescopes and reduces to an expectation over initial states:
\begin{equation}\label{equ:valuedice-objective}
\max_{\nu:\mathcal{S}\times\mathcal{A}\rightarrow \mathbb{R}} -\log \mathbb{E}_{(s, a)\sim d_E(s, a)}\big[ e^{\nu(s, a) - \mathcal{B}\nu(s, a)} \big] + (1-\gamma)\mathbb{E}_{s_0\sim p_0, a_0\sim\pi(\cdot|s_0)}\big[ \nu(s_0, a_0) \big].
\end{equation}
In the practical algorithm for a limited number of demonstrations, ValueDICE considers an alternative objective, with a controllable regularization based on experiences in the replay buffer: 
\begin{align}\label{equ:valuedice-practical-objective}
\min_{\pi}\max_{\nu:\mathcal{S}\times\mathcal{A}\rightarrow \mathbb{R}} -\log \mathbb{E}_{(s, a)\sim d_{\text{mix}}(s, a)}\big[ e^{\nu(s, a) - \mathcal{B}\nu(s, a)} \big] &+ (1-\alpha)(1-\gamma)\mathbb{E}_{s_0\sim p_0, a_0\sim\pi(\cdot|s_0)}\big[ \nu(s_0, a_0) \big] \nonumber \\
&+ \alpha \mathbb{E}_{(s,a)\sim d_{\text{RB}}}\big[ \nu(s, a) - \mathcal{B}^{\pi}\nu(s, a) \big],
\end{align}
where $d_{\text{mix}}(s, a)$ is the mixed distribution $d_{\text{mix}}(s, a) = (1-\alpha)d_{E} + \alpha d_{\text{RB}}$ and $\alpha$ is the controllable regularizer. 
The objective in Equation~\eqref{equ:valuedice-practical-objective} is shown to be equivalent to the KL-divergence between two mixed distributions, $\mathcal{D}_{\text{KL}}\big( (1-\alpha)d_\pi + \alpha d_{\text{RB}} || (1-\alpha)d_E + \alpha d_{\text{RB}} \big)$.

\section{Problems with ValueDICE}
In this section, we identify three problems with ValueDICE.

\subsection{Logorithms and exponentials of expectations}
As pointed out in the original paper~\cite{kostrikov2019imitation}, the objective~\eqref{equ:valuedice-practical-objective} of ValueDICE has two expectations that cannot be estimated without bias from mini-batches of samples.
To be specific, the first expectation has a logarithm applied to it, which makes mini-batch estimate of the gradient of this expectation biased.
The second expectation over the environment transition $p(\cdot|s, a,)$ computes $\mathcal{B}^{\pi}\nu(s, a)$ and has a log-exp applied to it, so its mini-batch approximated gradient estimate is also biased in general. 
These two problematic expectation terms stem from the use of KL-divergence and its special representation. 
In the original paper, this divergence is used to draw connections between reinforcement learning methods and distribution matching approaches.
Namely, the density ratios $\frac{d(s, a)}{d^*(s, a)}$ can be interpreted as the reward signals 
and the distribution matching is thus to maximize the sum of all density ratios between two mixed distribution~\cite{kostrikov2019imitation}. 
We show below that, instead of using a divergence, using another measure leads to an objective that can be more easily optimized and actually generalizes the classic feature expectation matching approaches~\cite{ratliff2006maximum,abbeel2004apprenticeship,ziebart2008maximum}.

\subsection{Mixture of samples}
The ValueDICE paper also points out that the number of expert samples may be small and lack diversity, which could potentially hamper policy learning. 
ValueDICE thus considers mixing expert demonstrations with off-policy samples and instead minimizes the distance between two mixed distributions, i.e., $\mathcal{D}_{\text{KL}}\big( (1-\alpha)d_\pi + \alpha d_{\text{RB}} || (1-\alpha)d_E + \alpha d_{\text{RB}} \big)$~\cite{kostrikov2019imitation}.
Therefore, there is no guarantee that the minimum ValueDICE converges to is the expert distribution.
Indeed, according to the convexity of the KL-divergence, 
$\mathcal{D}_{\text{KL}}\big( (1-\alpha)d_{\pi} + \alpha d_{\text{RB}} || (1-\alpha)d_E + \alpha d_{\text{RB}}  \big) \leq (1-\alpha) \mathcal{D}_{\text{KL}}( d_{\pi} || d_E)$, 
the objective in ValueDICE is only a lower bound of the $\mathcal{D}_{\text{KL}}(d_\pi || d_E)$. 
Thus, minimizing objective~\eqref{equ:valuedice-practical-objective} is weaker than minimizing objective~\eqref{equ:valuedice-kl-divergence},
and does not guarantee that the learned policy recovers the expert policy. 
Furthermore, the mixture of samples may not improve the overall performance of ValueDICE.
We show below that we can draw insights from feature expectation matching approaches to develop a more effective way to mitigate the lack of diversity in the demonstrations.

\subsection{Change of variable}
While the change of variable in ValueDICE simplifies the objective~\eqref{equ:valuedice-kl-divergence}, 
we show below that it implicitly assumes that the Markov chain induced by the policy $\pi$ is irreducible and aperiodic, and the state-action distribution is stationary.
In off-policy policy evaluation, for which such change of variable was originally proposed~\cite{nachum2019dualdice}, 
this assumption holds since the samples used for off-policy evaluation are from unknown but stationary distributions. 
However, in the policy training of imitation learning when the environment can be reset whenever a done signal is triggered by some terminal states, the induced Markov chain is not irreducible. 
As a result, $\mathbb{E}_{(s, a)\sim d_{\pi}} [\nu(s, a) - \mathcal{B}^{\pi} \nu(s, a)]$ is not equivalent to $(1-\gamma) \mathbb{E}_{(s, a)\sim d_0}[\nu(s, a)]$.
In the practical implementation, ValueDICE defined those terminal states as the \textit{absorbing state}, which has an extra dimension being set to one and all other dimensions being set to zero, similarly as in~\cite{kostrikov2018discriminator}. 
% Even though ValueDICE handles absorbing states of the environments as in~\cite{kostrikov2018discriminator}, 
Such special handling does not address the issue caused by the change of variable because the bias is introduced by the value of $\nu$ at those terminal states, not the states themselves, as we show in Section~\ref{section:avoid-bias}.

\section{Theoretical Analysis and SoftDICE}

To ease further analysis, we reformulate the affine constraints of Equation~\eqref{equ:feasible-constraint}. 
By multiplying both sides by $\pi(a^\prime|s^\prime)$, we have 
\begin{equation}\label{equ:stationarity-constraint}
d(s^\prime, a^\prime) = (1-\gamma) d_{0}(s^\prime, a^\prime) + \gamma \int_{s, a} \pi(a^\prime|s^\prime) p(s^\prime|s, a) d(s, a) \text{d}s \text{d}a \quad \forall s^\prime, a^\prime, 
\end{equation}
where $d_0(s, a) = p_0(s)\pi(a|s)$. 
Imitation learning can then be formulated as distribution matching with the above constraint~\cite{ho2016generative}: 
\begin{align}\label{equ:distribution-formulation}
\min_{\pi} \quad &  \mathcal{D} \big(d_{\pi}(s, a), d_E(s, a)\big), \nonumber \\
\text{subject to } \quad & d_{\pi}(s^\prime, a^\prime) = (1-\gamma) d_{0}(s^\prime, a^\prime) + \gamma \int_{s, a} \pi(a^\prime|s^\prime) p(s^\prime|s, a) d_{\pi}(s, a) \text{d}s \text{d}a \quad \forall s^\prime, a^\prime, 
\end{align}
where $\mathcal{D}$ is some distribution measure, e.g., KL or Jensen-Shannon divergence. 
The stationarity constraint comes from Proposition~\ref{prop:stationarity-constraint} and describes that the Markov chain induced by the policy $\pi$ should always be stationary. 
In ValueDICE the use of KL-divergence introduces the logarithms and exponentials of expectations. 
We can of course use other divergences, as suggested by~\cite{ke2019imitation,ghasemipour2020divergence},  but we show below that alternatively we can use one statistical measure (not a divergence), which yields a more feasible objective.  

\subsection{To remove logorithms and expontentials of expectations}
Inspired by the Wasserstein GAN~\cite{arjovsky2017wasserstein}, we instead adopt the Earth-Mover distance (EMD)~\cite{arjovsky2017wasserstein} and consider its Kantorovich-Rubinstein dual form~\cite{villani2008optimal}:
\begin{equation}\label{equ:new-objective}
\min_{\pi}\mathcal{D}(d_{\pi}, d_E) = \min_{\pi}\max_{||f||_{L}\leq 1} \mathbb{E}_{(s, a)\sim d_E}[f(s, a)] - \mathbb{E}_{(s, a)\sim d_\pi}[f(s, a)], 
\end{equation}
where the maximization is over all the 1-Lipschitz functions $f: \mathcal{X}\rightarrow\mathbb{R}$. 
If we replace $||f||_{L}\leq 1$ for $||f||_{K}$ (consider $K$-Lipschitz for some constant $K$), then we end up with $K\cdot \mathcal{D}(d_{\pi}, d_E)$. 
Compared to the ValueDICE objective, \eqref{equ:new-objective} has no exponential and logarithmic terms, and can be optimized with any stochastic gradient descent methods, e.g., Adam~\cite{kingma2014adam}.

\subsection{To combat the sparsity of demonstrations}
Before introducing our method, we first give intuition to help understand the above objective. 
Expanding the expectation over the policy, we have
$\mathbb{E}_{(s, a)\sim d_{\pi}}[f(s, a)] = \mathbb{E}^{\pi}\big[\sum_{t=0}^{\infty}\gamma^t f(s, a) | S_t=s, A_t=a \big]. $
If we interpret $f(s, a)$ as a feature for $(s, a)$, the objective in~\eqref{equ:new-objective} generalizes the feature expectation matching approaches in Apprenticeship Learning~\cite{abbeel2004apprenticeship}.
Specifically, denote the feature expectations as $\mu(\pi, f) = \mathbb{E}_{(s, a)\sim d_{\pi}}[f(s, a)] $, Equation~\eqref{equ:new-objective} can be re-written as 
\begin{equation}\label{equ:generalized-feature-matching}
\min_{\pi} \max_{||f|_L \leq 1}\Big[ \mu(\pi_E, f) - \mu(\pi, f) \Big]. 
\end{equation}
Similar to apprenticeship learning, distribution matching with this formulation in effect reduces imitation learning to finding a policy $\pi$ that induces feature expectations $\mu(\pi, f)$ close to $\mu(\pi_E, f)$\cite{syed2008game}.
The difference is that distribution matching generalizes such feature expectation matching to be over \emph{all} Lipschitz-continuous features $f\in\{f: ||f||_L\leq 1 \}$. 
% In this sense, the two-game playing \sw{what does this mean?} apprenticeship learning~\cite{syed2008game} is a special case in which only a fixed set of features is considered. 

In reality, the expert distribution $d_E$ is provided only as a finite set of samples, so in large environments, much of the state-action space is not visited, and exact feature expectation matching forces the learned policy to never visit those unseen state-action pairs simply due to lack of data~\cite{ho2016generative,kostrikov2019imitation}.
Inspired by the maximum entropy inverse reinforcement learning~\cite{ziebart2008maximum}, 
we employ the principle of maximum entropy to regularize the above optimization. Intuitively, the principle of maximum entropy~\cite{jaynes1957information}  prescribes the use of ``the least committed'' probability distribution that is consistent with known problem constraints~\cite{ziebart2010modeling}.
Specifically, we augment the objective function with an entropy regularizer $\mathbb{E}_{s\sim\pi_E}\big[ \mathcal{H}(\pi(\cdot|s)) \big]$
and learn a policy that maximizes the entropy over the expert state distribution.. This entropy term is defined differently from generative adversarial imitation learning (GAIL)~\cite{ho2016generative} in that the entropy term used here is averaged over the \textit{expert state distribution} $s\sim\pi_E$, while in GAIL the entropy regularizer  averages over the \textit{state distribution} induced by the policy $\mathbb{E}_{s\sim\pi}\big[ \mathcal{H}(\pi(\cdot|s)) \big]$. 
The reason for this is made clear below but we argue that such entropy definition still favors stochastic policies and resolves the ambiguity of policies with respect to the expert state distributions~\cite{ziebart2008maximum,haarnoja2018soft}. 
Accordingly, the new objective we consider is as follows
\begin{equation}\label{equ:min-max-objective}
\min_{\pi} \max_{||f||_{L}\leq 1} \Big[ \mathbb{E}_{(s, a)\sim d_E}[f(s, a)] - \mathbb{E}_{(s, a)\sim d_{\pi}}[f(s, a)] \Big]- \beta \mathbb{E}_{s\sim\pi_E}\big[ \mathcal{H}(\pi(\cdot|s)) \big],
\end{equation}
where the entropy coefficient $\beta$ determines the relative importance of the entropy term against the distribution matching objective.

\subsection{To avoid bias in off-policy formulation}\label{section:avoid-bias}
One can of course adopt the same re-parametrization trick used in ValueDICE~\cite{kostrikov2019imitation} and DualDICE~\cite{nachum2019dualdice} to turn on-policy term $\mathbb{E}_{(s, a)\sim d_{\pi}}\big[ f(s, a)\big]$ into off-policy. 
As we show below, this re-parametrization implicitly assumes that the Markov chain induced by the policy $\pi$ is ergodic during training. 
Namely, it equivalently states the stationarity constraint in~\eqref{equ:stationarity-constraint}:
as~\eqref{equ:stationarity-constraint} holds for all $(s, a)\in\mathcal{S}\times\mathcal{A}$, multiply right hand side (RHS) and left hand side (LHS) by a Lipschitz-continuous function $f(s, a)$, 
% \sw{I can't parse this sentence.}
\begin{equation*}
f(s^\prime, a^\prime) d_{\pi}(s^\prime, a^\prime) = (1-\gamma) f(s^\prime, a^\prime) d_{0}(s^\prime, a^\prime) + \gamma \int_{s, a} f(s^\prime, a^\prime) \pi(a^\prime|s^\prime) p(s^\prime|s, a) d_{\pi}(s, a) \text{d}s \text{d}a,
\end{equation*}
which holds for $\forall (s^\prime, a^\prime)\in \mathcal{S}\times\mathcal{A}$ and any $f$.
Taking the integral with respect to $s^\prime$ and $a^\prime$, we have
\begin{equation*}
\mathbb{E}_{(s, a)\sim d_{\pi}} [f(s, a)] = (1-\gamma) \mathbb{E}_{(s, a)\sim d_0}[f(s, a)] + \gamma \mathbb{E}_{(s, a)\sim d_{\pi}(s, a), s^\prime\sim p(\cdot|s, a), a^\prime\sim\pi(\cdot|s^\prime)}[f(s^\prime, a^\prime)]. 
\end{equation*}
Moving the second term on the RHS to the LHS and merging it into one expectation, we have
\begin{equation*}
\mathbb{E}_{(s, a)\sim d_{\pi}} [f(s, a) - \mathcal{B}_{\pi} f(s, a)] = (1-\gamma) \mathbb{E}_{(s, a)\sim d_0}[f(s, a)], 
\end{equation*}
where $\mathcal{B}^{\pi}$ is exactly what was defined in the change of variable trick in~\eqref{equ:change-of-variable-trick}.
Given that $f(s, a)$ is a Lipschitz-continuous function, $f(s, a) - \mathcal{B}^{\pi} f(s, a)$ is also Lipschitz-continuous.
The stationarity constraint is subsumed into the change of variable. 
However, in practice, the training environment may need to reset whenever some fatal actions are taken or adverse states are encountered, which breaks the original Markov chain and the samples generated by rolling out the policy are not from the stationary state-action distribution but instead from state marginal distribution~\cite{lee2019efficient}, i.e., the state-action distribution under a finite horizon, and the induced Markov chain is not irreducible, i.e., not ergodic. 
Specifically,
\begin{align*}
& \mathbb{E}_{(s, a)\sim d_\pi}\big[f(s, a) - \gamma \mathbb{E}_{s^\prime\sim P, a^\prime\sim\pi(s^\prime)}[f(s^\prime, a^\prime)] \big] \\
&= (1-\gamma)\sum_{t=0}^{T-1}\gamma^t \mathbb{E}_{s\sim p_t, a\sim\pi(s)}\big[ f(s, a) - \gamma\mathbb{E}_{s^\prime\sim P, a^\prime\sim\pi(s^\prime)}[f(s^\prime, a^\prime)] \big] \\
&= (1-\gamma) \sum_{t=0}^{T-1}\gamma^t\mathbb{E}_{s\sim p_t, a\sim\pi(s)}[f(s, a)] - (1-\gamma)\sum_{t=0}^{T-1}\gamma^{t+1}\mathbb{E}_{s\sim p_t, a\sim\pi(s)}[f(s, a)]\\
&= (1-\gamma)\mathbb{E}_{s\sim p_0, a\sim\pi(s)}[f(s, a)] - (1-\gamma)\gamma^T\mathbb{E}_{s\sim p_T, a\sim\pi(s)}[f(s, a)], 
\end{align*}
where $T$ is the horizon, which is always finite in training. 
Thus the off-policy term is not equivalent to the on-policy term as it introduces an extra bias term $(1-\gamma)\gamma^T\mathbb{E}_{s\sim p_T, a\sim\pi(s)}[f(s, a)]$. 
Also, it is clear that the bias is introduced by the value of $f$ at timestep $T$, not the terminal state-action pair. 
Thus, even with the use of absorbing state, i.e., setting a special flag for the terminal states~\cite{kostrikov2018discriminator,kostrikov2019imitation}, $f(s, a)$ could still be non-zero and thus the bias term remains. 
To avoid such bias, we propose to adopt  the unbiased off-policy Bellman residuals without rewards~\cite{baird1995residual}: $f(s, a) -  \mathbb{E}_{s^\prime\sim P, a^\prime\sim\pi(s^\prime)}[\gamma(1 - e) f(s, a)]$, where $e_t=1$ denotes the end of a finite episode at timestep $t$. 
The new objective is as follows:
\begin{equation*}
\min_{\pi} \max_{||f||_{L}\leq 1} \mathbb{E}_{(s, a)\sim d_E}\big[f(s, a)- \mathcal{B}^{\pi}_{e}f(s, a) \big] 
- (1-\gamma)\mathbb{E}_{s\sim p_0, a\sim \pi(\cdot|s)}[f(s, a)]
- \beta \mathbb{E}_{s\sim\pi_E}[\mathcal{H}(\pi(\cdot|s))], 
\end{equation*}
where $\mathcal{B}^{\pi}_{e}f(s, a) = \mathbb{E}_{s^\prime\sim P, a^\prime\sim\pi(s^\prime)}[(1 - e) f(s, a)]$. 
Similarly to ValueDICE, the $f$ resembles a ``value-function'' expressed in an off-policy manner, with expectations over expert demonstrations $d_E$ and the initial state distribution $p_0$.

\subsection{Offline imitation algorithm: SoftDICE}
One can also interpret the above optimization as policy iteration that alternates between policy evaluation, i.e., $\max_{||f||_{L}\leq 1}$, and policy improvement, i.e., $\min_{\pi}$, in the maximum entropy framework. 
In the policy evaluation step, we compute the value $f$ of a policy $\pi$ according to the entropy regularizer.
For a fixed policy $\pi$, the entropy term can be subsumed into $\mathcal{B}^\pi f(s, a)$, yielding a soft policy evaluation~\cite{fox2015taming,haarnoja2018soft}. 
Specifically, since the entropy is over the expert state distribution, we can subsume it into $\mathcal{B}^{\pi}_{e}f(s, a)$ by redefining the expectation as follows:
\begin{equation}\label{equ:new-bellman-with-entropy}
\mathcal{B}^{\pi}_{\mathcal{H}}f(s, a) = \mathbb{E}_{s^\prime\sim p(\cdot|s, a), a^\prime\sim\pi(\cdot|s^\prime)}\big[ \gamma(1-e) f(s, a) - \beta \log(\pi(a^\prime|s^\prime)) \big] \big]. 
\end{equation}
Therefore, the final objective is:
\begin{equation}\label{equ:new-distribution-formulation}
\min_{\pi} \max_{||f||_{L}\leq 1} \mathbb{E}_{(s, a)\sim d_E}\big[f(s, a)-\mathcal{B}^{\pi}_{\mathcal{H}}f(s, a)\big] - (1-\gamma)\mathbb{E}_{s\sim p_0, a\sim \pi(\cdot|s)}[f(s, a)].
\end{equation}
This new formulation echos the definition of the entropy regularizer, which is averaged over the expert state distribution.

To enforce the Lipschitz constraint on $f$, we use the gradient penalty from~\cite{gulrajani2017improved}.
Namely, a differentiable function is $1$-Lipschitz if and only if it has gradients with norm at most 1 everywhere. 
We thus add a gradient penalty regularizer with respect to samples from expert demonstrations~\cite{gulrajani2017improved} to the objective in~\eqref{equ:new-distribution-formulation} and rewrite as follows:
\begin{align}\label{equ:final-soft-dice-objective}
\min_{\pi} \max_{f} \Big[ \mathbb{E}_{(s, a)\sim d_E}\big[f(s, a)-\mathcal{B}^{\pi}_{\mathcal{H}}f(s, a)\big] &- (1-\gamma)\mathbb{E}_{s\sim p_0, a\sim \pi(\cdot|s)}[f(s, a)] \nonumber \\
& + \lambda \mathbb{E}_{(s, a)\sim d_E}\big[ ( ||\nabla f(s, a)||_2 - 1)^2 \big] \Big].
\end{align}
% We also use the orthogonal regularization for $\pi$ as used in the ValueDICE since the orthogonality of weights can stabilize the training~\cite{bansal2018can}. 

In terms of the initial state sampling $s\sim p_0$, we adopt the scheme used in ValueDICE to treat the expert demonstrations as samples from $p_0$. 
The underlying intuition is to consider each full trajectory $(s_0, a_0, s_1, a_1, ..., s_T)$ as $T$ distinct virtual trajectories $\{ (s_t, a_t, s_{t+1}, a_{t+1}, ..., s_T \}$. 
This sampling scheme does not affect the optimality of the learned policy $\pi$ since in Markov environments an expert policy is expert regardless of the initial state distribution~\cite{puterman2014markov}.

\begin{algorithm}
\caption{SoftDICE}
\begin{algorithmic}[1] %enables line numbers
\STATE Initialize replay buffer $\mathcal{R}_{E}$ with expert demonstrations $\{ (s_E, a_E, s^\prime_E, e) \}$
\FOR{$t=1$ to $T$}
\STATE Sample $\{(s_E^{(i)}, a_E^{(i)}, s^{\prime(i)}_E, e) \}_B\sim\mathcal{R}_E$.
\STATE Obtain $a^{\prime(i)}\sim\pi_\theta(\cdot|s^{\prime(i)})$, $a^{\prime(i)}_E\sim\pi_\theta(\cdot|s_E^{\prime(i)})$, for $i=1, ..., B$
\STATE Compute loss on expert samples:
\STATE $\qquad J_{E} = \frac{1}{N}\sum_{i=1}^{N}\big[ f_\phi\big(s_E^{(i)}, a_E^{(i)} \big) - \gamma (1-e) f_\phi\big(s_E^{\prime(i)}, a_E^{\prime(i)}\big) + \beta \log\pi(a_E^{\prime(i)}|s_E^{\prime(i)}) \big]$
\STATE Sample $\{s_0^{(i)}\}_B\sim\mathcal{R}_E$.
\STATE Obtain $a^{(i)}_0\sim\pi_\theta(\cdot|s^{(i)}_0)$, for $i=1, ..., B$
\STATE Compute loss on initial samples:
\STATE $\qquad J_{\pi} = (1-\gamma) \frac{1}{N}\sum_{i=1}^{N}\big[f_\phi\big(s_0^{(i)}, a_0^{(i)}\big)\big]$
\STATE Compute gradient penalty on expert samples:
\STATE $\qquad J_{\text{GP}} = \frac{1}{N}\sum_{i=1}^{N}\big[||\nabla f_\phi\big(s_E^{(i)}, a_E^{(i)}\big)||_2 - 1\big]^2$
\STATE Update $\theta\leftarrow\theta - \eta_{\theta}\nabla_{\theta}(J_{E} - J_{\pi})$
\STATE Update $\phi\leftarrow\phi + \eta_{\phi}\nabla_{\phi}(J_{E} - J_{\pi} + J_{\text{GP}})$
\ENDFOR
\RETURN $\pi_\theta$
\end{algorithmic}
\label{algo:soft-dice}
\end{algorithm}

Our formulation is completely off-policy and can thus be applied for offline imitation learning. 
As the ``value function'' $f$ is updated in a soft manner, we name the resulting algorithm \textit{SoftDICE}. 
We show in the next section that, unlike ValueDICE, SoftDICE does not need to mix the expert demonstrations with replay buffer samples. 
We use a similar optimization scheme in ValueDICE to alternate the parameter update of $f$ and $\pi$. 
The SoftDICE algorithm is presented in Algorithm~\ref{algo:soft-dice} and runs as follows:
It first samples a batch of expert demonstration pairs from the expert replay buffer and computes the loss $J_{E}$ on expert samples. 
Then it conducts the same sampling from the expert replay buffer and computes the loss $J_{\pi}$ on initial samples. 
The gradient penalty is also computed with respect to the expert samples. 
SoftDICE then updates parameters $\phi$ and $\theta$ with different losses according to~\eqref{equ:final-soft-dice-objective} using a stochastic gradient optimizer. 
This procedure repeats for $T$ iterations and returns the learned policy.

\section{Experiments}

\subsection{Mujoco benchmark tasks}

We evaluate the performance of SoftDICE on the suite of MuJoCo continuous control~\cite{todorov2012mujoco}, interfaced through OpenAI Gym~\cite{brockman2016openai}. 
The MuJoCo control tasks are characterized by varying difficulties and have been widely used for benchmarking imitation learning algorithms for continuous action space.
Following the ValueDICE implementation, 
all algorithms use networks with an MLP architecture with 2 hidden layers and 256 hidden units. 
All networks use orthogonal initialization. 
We use the Adam optimizer with learning rate $10^{-3}$ for the Lipschitz continuous function $f$, and $10^{-5}$ for the policy $\pi$. 
We use gradient penalties from~\cite{gulrajani2017improved} to enforce Lipschitz continuity and regularize the actor network with orthogonal regularization~\cite{brockman2016openai}.
The entropy coefficient $\beta$ is tuned to be $0.01$ for best performance on all tasks except Ant, which is set to $0.2$. 
We repeat each experiment with 5 random seeds and report means and standard deviations in all plots.

\begin{figure}[ht]
    \centering
    \includegraphics[width=0.95\textwidth]{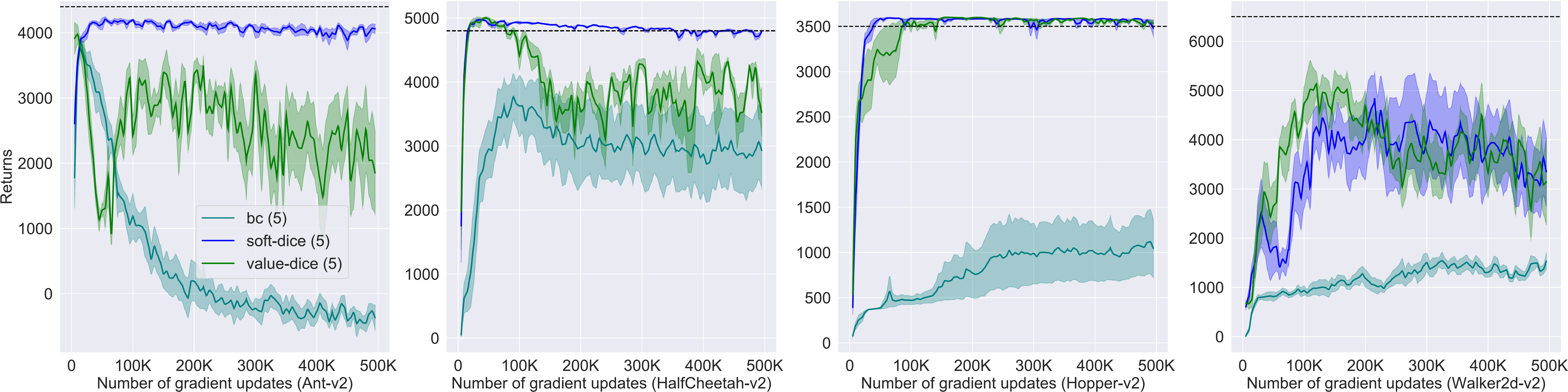}
    \caption{Offline evaluation with only 1 demonstration trajectory (no sampling and policy rollout; dash lines mark the expert performance).}
    \label{fig:offline-comparison}
\end{figure}

Since SoftDICE operates completely offline, we first compare it with two offline imitation learning methods: behavioral cloning (BC)~\cite{pomerleau1991efficient} and the offline variant of ValueDICE. 
Theoretically, ValueDICE considers a completely off-policy objective and can thus be adapted to be offline by setting the ``replay buffer regularization'' coefficient to zero~\cite{kostrikov2019imitation}. 
This means that no off-policy samples are mixed into the expert demonstrations and used to regularize training. 
We use the open source implementation without changing any hyperparameters.\footnote{\url{https://github.com/google-research/google-research/tree/master/value\_dice}, under Apache License 2.0.}	
During training, we randomly sample only one full trajectory from the demonstrations provided by the code repository.
Results are presented in Figure~\ref{fig:offline-comparison}. 
With such limited data, BC completely fails to recover any expert policies in all benchmark tasks. 
However, SoftDICE quickly learns the optimal policy in Ant-v2, HalfCheetah-v2 and Hopper-v2 in less than 50K gradient updates.
Notably, as training proceeds in Ant-v2 and HalfCheetah-v2, the performance of SoftDICE still remains close to the expert level while the performance of ValueDICE drops significantly, which could be potentially caused by the biased gradient updates. 
We also notice that all algorithms fail to achieve the expert performance on Walker2d-v2. 
This could be that, with a single trajectory, learning the expert policy for this specific environment may be very challenging. 
We provide a full comparison across a varying number of trajectories in the appendix and SoftDICE can achieve the expert performance on Walker2d-v2 when the number of trajectories increases.

\begin{figure}
    \centering
    \includegraphics[width=0.95\textwidth]{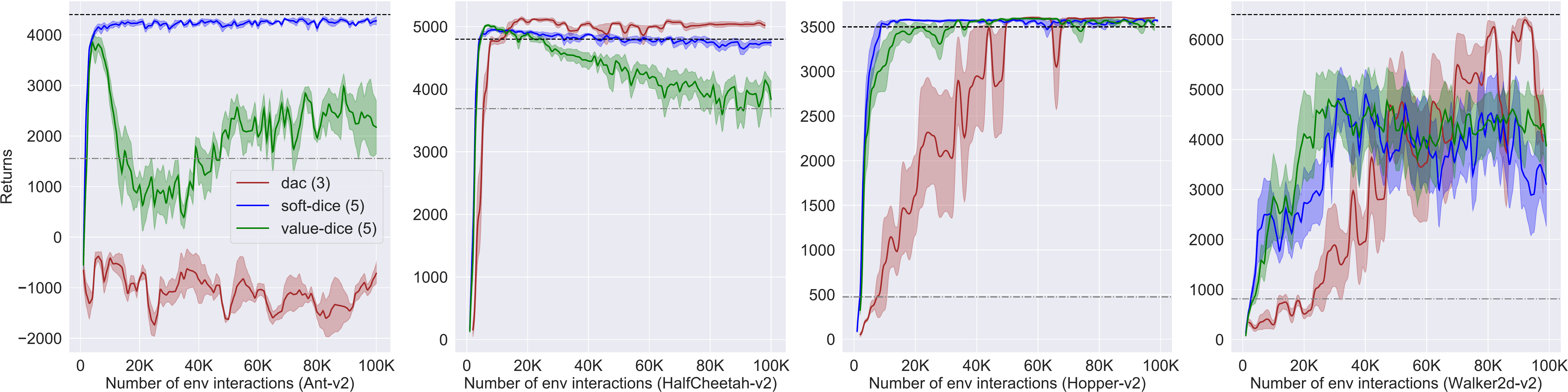}
    \caption{Evaluation with only 1 demonstration trajectory and samples from policy rollout (black dash lines and grey dash lines mark the expert performance and BC performance respectively).}
    \label{fig:online-comparison}
\end{figure}

The original ValueDICE needs to roll out the policy in training to collect samples to increase the diversity of the expert demonstrations. 
The original paper reported that, by ``regularizing'' expert demonstrations with off-policy samples, ValueDICE could be enhanced for a small number of expert samples. 
We thus adopt the same enhancement for SoftDICE and also compare it against ValueDICE in the online setting, where the policy is allowed to roll out for collecting samples.  
We also consider Discriminator-Actor-Critic (DAC)~\cite{kostrikov2018discriminator} as a baseline since DAC is an adversarial imitation approach with good sample efficiency. 
Similarly to the offline setting, we also randomly sample one trajectory as the expert demonstration. 
We roll out the policy after every 5 gradient updates to the policy and use the same small replay buffer regularizer $\alpha=0.1$ as in ValueDICE. 
Figure~\ref{fig:online-comparison} shows that the policy rollout improves the performance of ValueDICE in general. 
However, SoftDICE still outperforms ValueDICE in Ant-v2 and HalfCheetah-v2. 
SoftDICE also achieves better sample efficiency than DAC in Ant-v2 and Hopper-v2. 
We also notice the same poor performance of all algorithms on Walker2d-v2, which, we argue, could be caused by the scarcity of demonstration data in training. 

\subsection{Ablation study on entropy regularization}
\begin{wrapfigure}{r}{0.4\textwidth}
  \begin{center}
    \includegraphics[width=1.0\linewidth]{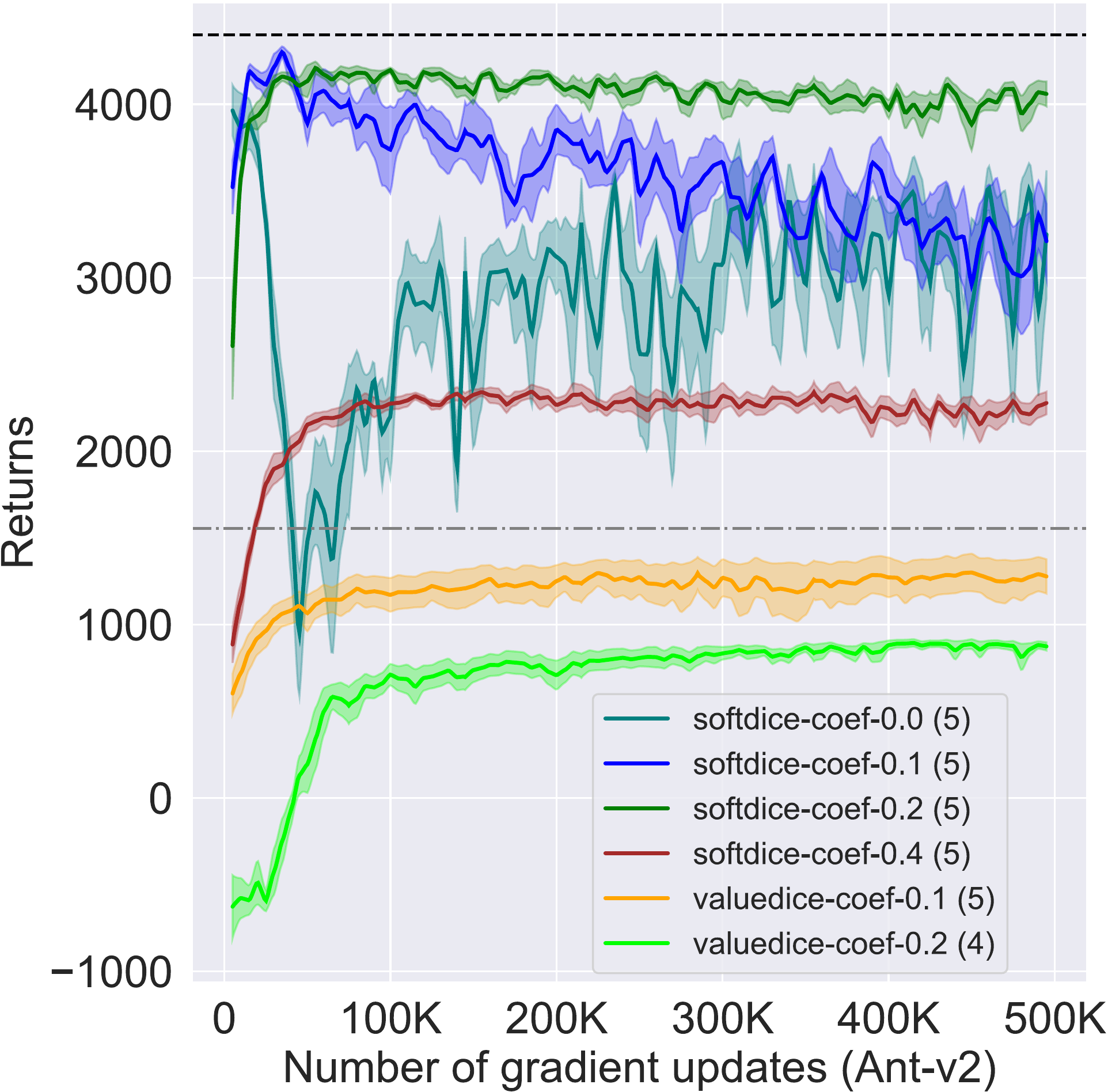}
  \end{center}
  \caption{Ablate entropy regularization}
  \label{fig:ablation-entropy}
\end{wrapfigure}

One key difference between SoftDICE and ValueDICE is that SoftDICE uses maximum entropy to regularize the policy training. 
We therefore ablate this entropy regularization on both methods. Specifically, we consider using different entropy coefficients in the offline training for SoftDICE,
and also add an additional entropy term for ValueDICE. 
Due to its formulation, ValueDICE cannot integrate such a policy entropy term into any of its expectations and we thus treat the entropy term as an extra loss to the policy optimization. 
We use  Ant-v2 as the ablation task as the performance gap between these methods is large. 
Figure~\ref{fig:ablation-entropy} shows that entropy regularization does have an impact on the convergence of SoftDICE: no entropy regularization results in a performance drop similar to that in ValueDICE and a small regularizer, e.g., $0.1$ and $0.2$ stablizes  training while a large regularizer impedes learning.
By contrast, even with the same levels of entropy regularization, ValueDICE still fails to perform comparably, which implies that entropy regularization itself is not the key for ValueDICE to learn good policies in this task.

\section{Related Work}

The study of imitation learning traces back to behavior cloning (BC)~\cite{pomerleau1991efficient}, in which the optimal policy is learned by minimizing the errors between policy predictions and human demonstrations. 
Such supervised methods suffer greatly from compounding errors caused by distribution shift when the number of given demonstrations is limited~\cite{ross2011reduction,ho2016generative}. 
Recent imitation learning approaches model the problem as a decision-making task based on a Markov decision process, 
which seeks to learn a reward function that could then be used as a proxy to learn the optimal policy, a.k.a.\ inverse reinforcement learning (IRL)~\cite{abbeel2004apprenticeship}. 
Many early IRL approaches decompose the reward as a linear combination of features and use the feature expectations of a policy as the proxy quantity to measure the similarity between expert policy and an arbitrary policy, i.e., feature expectation matching~\cite{abbeel2004apprenticeship, syed2008game,ziebart2008maximum}. 
Our formulation of off-policy distribution matching yields an intermediate objective that resembles the feature expectation matching. 
In particular, it is close to the game-theoretic approach~\cite{syed2008game}, which maximizes performance relative to the expert and also adversarially updates the feature combinations. 
The key difference is that our formulation considers a much larger feature space, \textit{all} Lipschitz continuous functions, while the game-theoretic approach constrains features to be in a limited set. 
In general, learning a reward function from a given set of demonstrations is ill posed, as there could be many reward functions that explain the demonstrations. 
Recent studies focus on adversarial imitation~\cite{ho2016generative} which defines imitation learning as a distribution matching problem and leverages GANs~\cite{goodfellow2014generative} to minimize the Jensen-Shannon divergence between distributions induced by the expert and the learning policy. 
This approach avoids the ambiguity of learning a reward function from demonstrations but is generally sample intensive.

To improve sample efficiency, many methods extend adversarial imitation to be off-policy. 
For instance, Discriminator-actor-critic (DAC)~\cite{kostrikov2018discriminator} improves the sample efficiency by reusing previous samples stored in a relay buffer.
However, this approach still relies on the non-stationary reward signals that are generated by the discriminator, which can make the critic estimation hard and training unstable. 
Recent work proposes to train a fixed reward function through estimating the support of demonstrations and then trains the critics with the fixed reward~\cite{wang2019random}.
This support estimation itself can be hard given that only a limited number of empirical samples are available from the considered distributions.
Another line of off-policy distribution matching approaches focuses on estimating the critics directly without learning any reinforcement signals~\cite{sasaki2018sample,kostrikov2019imitation}.
The state-of-the-art along this line is  ValueDICE~\cite{kostrikov2019imitation}, which casts distribution matching as off-policy density ratio estimation and updates the policy directly via a max-min optimization. 
However, as we show in the analysis and experiments, there are several problems in the formulation of ValueDICE that can significantly influence its empirical performance and convergence guarantees.

\section{Conclusion}
We present SoftDICE, an off-policy distribution matching approach for sample-efficient imitation learning. 
SoftDICE fixes several key problems in ValueDICE~\cite{kostrikov2019imitation}:
First, the objective of ValueDICE contains logarithms and exponentials of expectations, 
for which the mini-batch gradient estimate is always biased.
Second, the objective in ValueDICE is regularized with replay buffer samples when expert demonstrations are limited in number, 
which however changes the original distribution matching objective. 
Third, the re-parametrization trick used to derive the off-policy objective implicitly assume that the induced Markov in training is always ergodic, which rarely holds in practice. 
We leverage the Earth-Mover Distance and consider an entropy-regularized off-policy objective, which eliminates all the aforementioned issues in ValueDICE and also yields a completely offline algorithm. 
Empirical results show that SoftDICE recovers the expert policy with only one demonstration trajectory and no further on-policy/off-policy samples.
Also, it stably outperforms ValueDICE and other baselines in terms of sample efficiency on Mujoco benchmark tasks.

\textbf{Limitations.}
We see limitations in the following aspects. 
First, SoftDICE is not directly compared with other offline imitation algorithms. 
As SoftDICE is proposed to fix problems in ValueDICE, the comparison with other offline methods is left for future work.
Second, we only evaluate SoftDICE on the MuJoCo benchmark tasks. 
It would also be interesting to see how SoftDICE could be generalized to real-world applications, e.g., human-like autonomous driving. 

\textbf{Societal impact.}
The study of imitation learning could be important to understand human intelligence.
Sample efficiency, either the number of interactions required for training or the amount of demonstration data needed for imitation, could be the first step to understand how to learn fast and efficiently as human beings. 
SoftDICE learns a policy with only limited expert data and no further interactions, and provides a computational perspective to understand the efficiency in human imitation learning. 
However, SoftDICE could be misused to deceive people by creating fake behaviors, e.g., exactly imitating one's gaits. 
Also, if the demonstrations contain ethnic-sensitive human data, the policy learned via SoftDICE could be biased. 

\bibliographystyle{plain}
\bibliography{main}

\appendix

\newpage
\section{Appendices}

\subsection{Offline evaluation with different number of demonstration trajectories}

\begin{figure}[h]
    \centering
    \includegraphics[width=1.0\textwidth]{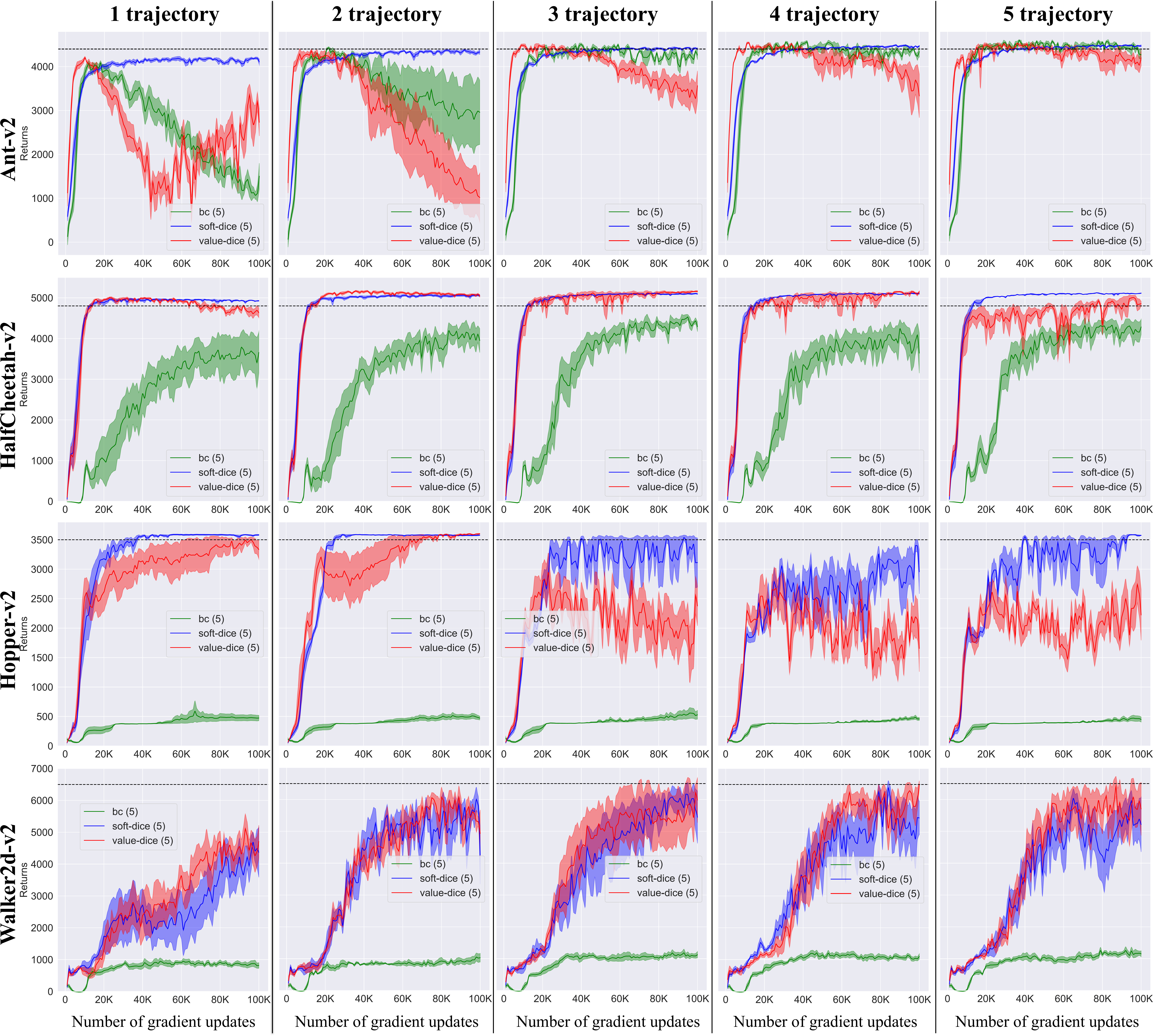}
    \caption{Offline evaluation with different number of demonstration trajectories (black dash lines mark the expert performance.}
    \label{fig:offline-comparison-all-trajs}
\end{figure}

\end{document}